\documentclass[letterpaper, 10 pt, conference]{ieeeconf}  
\pdfoutput=1 

\IEEEoverridecommandlockouts                              

\usepackage{caption}
\usepackage{subcaption}
\usepackage{graphics} 
\usepackage{epsfig} 
\usepackage{times} 
\usepackage{amsmath} 
\usepackage{amssymb}  
\usepackage{bm}
\usepackage{algorithm}
\usepackage{multirow}
\usepackage{verbatim}
\usepackage{graphicx}
\usepackage{hyperref}
\usepackage{graphicx,fancyhdr}
\usepackage{todonotes}
\usepackage{tabularx}
\usepackage{multirow}
\usepackage{multicol}
\usepackage{newtxmath,newtxtext}
\usepackage{booktabs} 
\usepackage{amsmath}

\usepackage{color}
\usepackage[export]{adjustbox}

\title{\LARGE \bf
How Can Everyday Users Efficiently Teach Robots by Demonstrations?
\vspace*{-4mm}
}

\author{Maram Sakr$^{*1,3}$, Zhikai Zhang$^{2}$, Benjamin Li$^{1}$,
\\Haomiao Zhang$^{1}$, H.F. Machiel Van der Loos$^{1}$, Dana Kuli{\'c}$^3$ and Elizabeth Croft$^{4}$
\thanks{$^{1}$Mechanical Engineering Department, University of British Columbia.}
\thanks{$^{2}$Mechanical Engineering Department, Carnegie Mellon University; Pittsburgh, USA.}
\thanks{$^{3}$Electrical and Computer Systems Engineering Department, Monash University.}
\thanks{$^{4}$Mechanical Engineering Department, University of Victoria}
\thanks{$^{*}$Email: maram.sakr@ubc.ca}
\vspace*{-7mm}
} 
\begin{document}

\maketitle
\thispagestyle{empty}
\pagestyle{empty}

\begin{abstract}
Learning from Demonstration (LfD) is a framework that allows lay users to easily program robots. However, the efficiency of robot learning and the robot’s ability to generalize to task variations hinges upon the quality and quantity of the provided demonstrations. Our objective is to guide human teachers to furnish more effective demonstrations, thus facilitating efficient robot learning. To achieve this, we propose to use a measure of uncertainty, namely task-related \textit{information entropy}, as a criterion for suggesting informative demonstration examples to human teachers to improve their teaching skills.  
In a conducted experiment (N=24), an augmented reality (AR)-based guidance system was employed to train novice users to produce additional demonstrations from areas with the highest entropy within the workspace. These novice users were trained for a few trials to teach the robot a generalizable task using a limited number of demonstrations. Subsequently, the users' performance after training was assessed first on the same task (retention) and then on a novel task (transfer) without guidance. The results indicated a substantial improvement in robot learning efficiency from the teacher's demonstrations, with an improvement of up to 198\% observed on the novel task. Furthermore, the proposed approach was compared to a state-of-the-art heuristic rule and found to improve robot learning efficiency by 210\% compared to the heuristic rule.

\end{abstract}

\section{INTRODUCTION}
        \label{sec:introduction}
        In learning from demonstration (LfD), robots are taught tasks through provided examples of how to perform them~\cite{argall2009survey}. This approach identifies promising search spaces for more efficient and effective robot learning, while also enabling users to convey task information naturally, as opposed to direct task programming~\cite{argall2009survey, billing2010formalism}. This natural interaction benefits novice users who lack the technical background for programming via an interface or deep understanding of the underlying system~\cite{chernova2014robot}.

LfD techniques build upon standard Machine Learning (ML) methods that have had great success in a wide range of applications~\cite{chernova2014robot}. However, learning from a human teacher poses additional challenges, such as limited human patience, low-quality and inefficient data~\cite{billard2016learning}. Collecting the training data set in any learning system is critical to a successful learning process. Training data must be representative of the states that the robot will encounter in the future. The size and diversity of the training data set will determine the speed and accuracy of learning, including its generalization characteristics. Moreover, data changes have a more significant impact on generalization performance than algorithm changes~\cite{deng2009imagenet}. Thus, this paper focuses on optimizing the input (i.e., demonstration data) to the learning algorithm, rather than altering the algorithm itself. 

Human teaching is often optimized to human learning and may not naturally suit the needs of arbitrary machine learners~\cite{cakmak2014eliciting}. Nevertheless, human teachers can adjust to a specific learner's requirements~\cite{khan2011humans}. To facilitate this adaptation, we introduce the concept of "teaching guidance", which consists of instructions provided to human teachers, with the goal of influencing their choice of examples towards those that are most informative for a particular learner.  

\begin{figure}[t]
\centering
\includegraphics[width=0.48\textwidth]{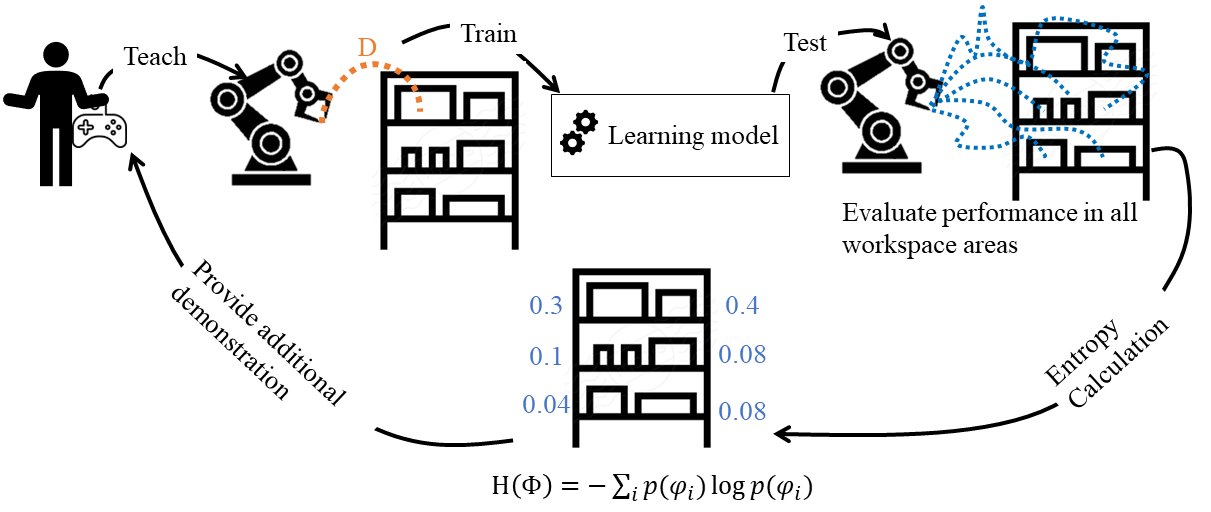}
\caption{System Overview: Using information entropy for guiding users to the highest uncertainty areas in LfD.}
\vspace*{-3mm}
\label{fig:overview}
\end{figure}

In this paper, we aim to help human teachers provide demonstrations efficiently by using information entropy as the criterion for selecting demonstrations for robot learning and generalization. Specifically, task-dependent information entropy is introduced in the LfD pipeline to define uncertainty across all task space areas as shown in Fig.~\ref{fig:overview}. Furthermore, we propose integrating Augmented Reality (AR) with the entropy-based guidance system as a training framework to help novice users learn to better identify informative demonstrations. We also compare our information entropy approach with a state-of-the-art heuristic rule~\cite{sena2020quantifying} for selecting demonstrations in the task space in terms of robot learning efficiency.

\section{RELATED WORK}
        \label{sec:related_works}
        Teaching robots through demonstration poses challenges for lay users as they may struggle to provide useful~\cite{9832744, khan2011humans} and sufficient demonstrations~\cite{sena2020quantifying}. Robots can become more proactive in their learning by requesting additional demonstrations~\cite{shon2007active} or seeking clarification~\cite{cakmak2012designing}. However, this shift to active learning can lead to users feeling frustrated and disengaged~\cite{cakmak2010designing}. Typically, learning from an expert teacher who selects examples optimally is faster than an active learning process where learners choose their own examples~\cite{goldman1995complexity}. Therefore, teachers require guidance in selecting optimal examples for robot learning and generalization.  

Whilst there is a great deal of research on policy-learning methods for LfD, there is considerably less research on the quality of teaching and how this might be improved~\cite{argall2009survey, billing2010formalism}. In fact, effective teaching can be challenging, even when instructing humans, requiring an understanding of the learner's knowledge, suitable teaching strategies, and more. Teaching robots presents an even greater challenge. Cakmak and Takayama~\cite{cakmak2014teaching} proposed using various instructional materials to guide novice teachers on how to teach robots using kinesthetic teaching and a spoken dialogue interface. Mohseni-Kabir et al.~\cite{mohseni2015interactive} designed an interactive system for teaching users a hierarchical task structure. Study participants were given explicit instructions on how to teach the robot for each task. Such an approach is untenable for scaling up to a vision of ubiquitous robotics, as it is impractical for experts to teach every end user how to program robots for every desired task. Instead, we examine how people teach novel tasks to robots in the absence of a roboticist’s explicit tutelage and focus on guiding users to optimally select a set of demonstrations for the robot to learn a task.  

Some researchers have attempted to address the problem of demonstration efficiency. Calinon and Billard~\cite{calinon2007incremental} and Weiss et al.~\cite{weiss2009teaching} both described an incremental learning approach where teachers decide on the next demonstration by observing the robot attempt the taught skill in new locations. This overlooks the question of \textit{how} the attempts should be selected to hasten robot learning progress. Sena et al.~\cite{sena2020quantifying} proposed heuristic rules for deciding demonstration locations in the task space. These rules are: (i) provide one demonstration, starting from anywhere in the task space, (ii) continue providing demonstrations within 4 cm of the first demonstration, until the first demonstration is surrounded by successful test points, (iii) provide further demonstrations within 4 cm of the successful test points, in the area with the greatest number of failed test points. These rules are limited to their experimental setup and are hard to generalize for a different task. To overcome this limitation and provide a more general framework, we propose using a task-independent rule (information entropy) to guide users to provide efficient demonstrations. 

A closely related field of research that considers optimal teaching of machine learning systems is \textit{machine teaching}. It designs the optimal training data to drive the learning algorithm to a target model~\cite{zhu2015machine}. The teaching dimension model represents the minimum number of instances a teacher must reveal to uniquely identify any target concept~\cite{goldman1995complexity} while the curriculum learning principle is a general approach in which training examples are presented to the learner in a sequence that is tailored to learning, progressing from simple to difficult concepts~\cite{bengio2009curriculum}. This opens the door for further research on which optimal teaching strategy should be used with robots. Khan et al.~\cite{khan2011humans} compared the teaching dimension model and curriculum learning principle in teaching a simple 1D classification problem, while Cakmak and Lopes~\cite{cakmak2012algorithmic} studied optimal teaching in sequential decision tasks. Other related approaches examine how a robot can give informative demonstrations to a human~\cite{huang2019enabling}, or formalize optimal teaching as a cooperative two-player Markov game~\cite{hadfield2016cooperative}; however, neither approach addresses the machine teaching problem of finding the minimum number, quality and locations of demonstrations needed to teach a task.

\section{PROPOSED APPROACH}
        \label{sec:approach}
        \begin{figure*}[t]
\centering
\includegraphics[width=0.90\textwidth]{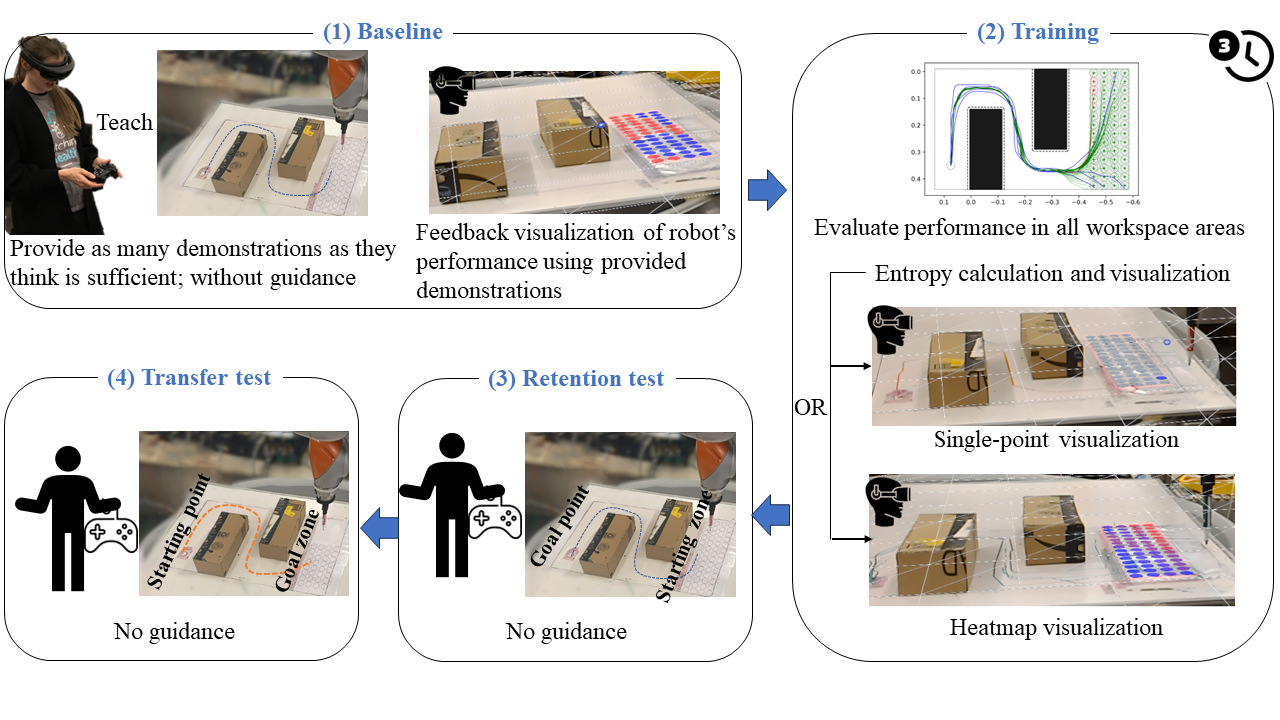}
\caption{Overview of the user study procedure. \textbf{(1) Baseline:} The user provides as many demonstrations as they think would be sufficient for the robot to learn the task. \textbf{(2) Training:} The user is guided to the uncertain areas for three trials. \textbf{(3) Retention test:} The user is evaluated on the same task without guidance. \textbf{(4) Transfer test:} The user teaches the robot to move from a specific starting point to a wider goal zone without guidance.}
\vspace*{-3mm}
\label{fig:training}
\end{figure*}

In this paper, we propose using information entropy as the criterion for suggesting to the user the location of the next demonstration. Entropy in information theory is a measure of the uncertainty or randomness of a system or signal~\cite{shannon2001mathematical}. If the entropy is higher, more information is needed to represent an event. Mathematically, entropy is represented by the Shannon entropy formula, which is given by:
\begin{equation}
    H(\Phi) = -\sum_i p(\phi) \log p(\phi)
    \label{equ:entropy}
\end{equation}

\noindent where $p(\phi)$ is the probability of a particular event $i$. This probability can be calculated as a weighted sum of different task-dependent features~\cite{ziebart2008maximum}.

Figure~\ref{fig:overview} provides an overview of the process of teaching a robot an example task, which involves retrieving items from a shelf using an efficient set of demonstrations that enable generalization across the robot's workspace. The task requires the robot to pick up all necessary items from a shelf without encountering collisions. To efficiently accomplish this task, first, the workspace and the success criteria for the task are defined. The workspace, in this context, refers to the shelf that the robot needs to access entirely. Three success criteria are considered as follows: a) the Robot’s end effector reaches the goal location, b) no collisions occur with obstacles in the environment, and c) all points on the trajectory are within the workspace (i.e., the robot should not deviate from its designated workspace). Next, a worker provides a demonstration of picking up an item from the shelf. Using the provided demonstration(s), the robot learns a model of how to perform the task. The robot is then evaluated on all regions of the shelf, and the entropy of each region is calculated as shown in equation~\ref{equ:entropy}. The probability is defined based on the success criteria as follows:
\begin{equation}
\label{equ:prob}
p(\phi) = \alpha d_{goal} + \beta n_{collision} + \gamma n_{outside}
\end{equation}
where $d_{goal}$ is the distance between the trajectory endpoint and the task goal point, $n_{collision}$ is the number of samples in collision with the obstacles, and $n_{outside}$ is the number of samples outside the task space. The coefficients $\alpha, \beta, \gamma$ are the weights for each feature of the probability. Each feature is normalized by its maximum value before using it in entropy $H(\Phi)$ calculations. Areas with high entropy are highlighted to guide the user in providing additional demonstrations. This iterative process continues until the entropy for all regions becomes sufficiently low, indicating that the robot is confident in generalizing across all regions in the workspace. Importantly, our information entropy approach is not influenced by the user's expertise level or the learning algorithm. Even if low-quality demonstrations are provided, the guidance system may recommend more demonstrations to facilitate effective robot learning and generalization.

To guide the users to the uncertain areas in the task space, a Microsoft HoloLens AR head-mounted display~\cite{kress201711} is used to visualize the entropy values in task space areas. Providing the user with more insight into the learning process may lead to more effective demonstrations, but providing too much information may also overwhelm and hinder the user, especially if they are untrained users. To find an adequate amount of information to be presented in the visual interface about the robot's learning progress, we proposed two visualization schemes. \textit{Heatmap visualization} that show the entropy values across the task space as a heatmap and \textit{single-point visualization} that only highlights the most uncertain area in the task space to provide an additional demonstration as shown in Fig.~\ref{fig:training}-2.

For task learning, we used the Task Parameterized Gaussian Mixture Model (TP-GMM)\cite{calinon2016tutorial}. TP-GMMs have been extensively used for LfD~\cite{pervez2018learning, osa2018algorithmic}, and they show good generalization using a limited set of demonstrations. TP-GMMs model a task with parameters that are defined by a sequence of coordinate frames. In a \(D\) dimensional space, each task parameter/coordinate frame is given by an $A \in \mathbb{R}^{D \times D}$ matrix indicating its orientation and a $b \in \mathbb{R}^D$ vector indicating its origin, relative to the global frame. A \(K\)-component mixture model is fitted to the data in each local frame of reference. Each GMM is described by $\left(\pi_k, \mathbf{\mu}^{(j)}_k, \mathbf{\Sigma}^{(j)}_k\right)$, referring to the prior probabilities, mean, and covariance matrices for each component \(k\) in frame \(j\), respectively.

To use the local models for trajectory generation, they must be projected back into the global frame of reference and then combined into one global model. Continuous trajectories can then be generated from the global mixture model using Gaussian Mixture Regression (GMR), Calinon \cite{calinon2016tutorial} provides further details. A Bayesian Information Criterion (BIC) was used to define the optimal number of \(K\)-Gaussian components to fit the demonstrations.


        
\section{EVALUATION}
        \label{sec:evaluation}
        To evaluate the proposed entropy system, a preliminary experiment in simulation comparing the proposed approach and one of the state-of-the-art heuristic guidance rules is presented. It aims to study the impact of the proposed approach on robot learning efficiency. After that, a user study is presented to explore the contribution of the proposed approach to improving human teaching skills. 

\subsection{Experimental Task}
\label{subsec:task}
\begin{figure}[t]
\centering
\includegraphics[width=0.40\textwidth]{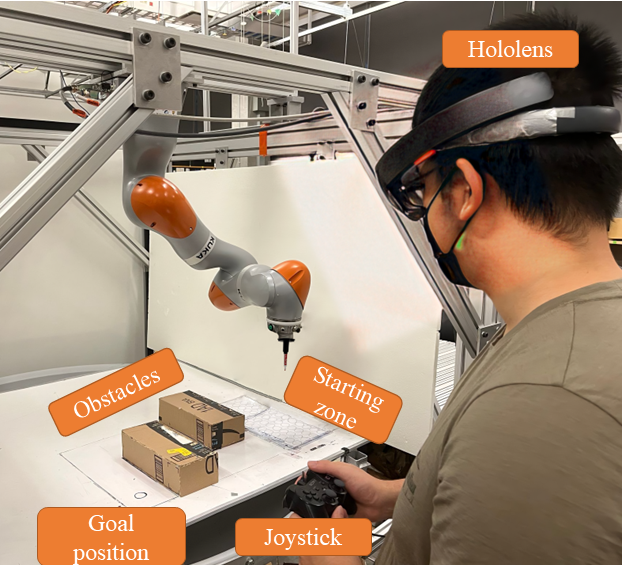}
\caption{Experimental setup used in the user study.}
\vspace*{-3mm}
\label{fig:exp_setup}
\end{figure}

A relatively simple task was chosen following~\cite{sena2020quantifying} to study the effectiveness of the proposed system on both robot learning efficiency and users' teaching performance. As shown in Fig.~\ref{fig:exp_setup}, the task is to teach the Kuka IIWA LBR14 robot to navigate a 2D maze from a starting zone to a target point. The robot should avoid collisions with the obstacles in the workspace and it should not go outside the workspace specified by a two-dimensional bounding rectangle (45 cm by 72 cm) drawn on the whiteboard. The starting zone is defined by a 45 cm by 15 cm rectangle. In the experiment, the starting zone is discretized to a 4 x 14 test grid with 56 circles separated by 0.5 cm. Participants use the joystick to teach the robot to navigate through the maze. The teaching process is finished if the robot can generate a "successful" trajectory for 90\% of the points in the starting zone to the goal position. The success criteria of the trajectories are similar to what is defined in the example task in Section~\ref{sec:approach}. Microsoft Hololens is used to overlay a geometrically accurate starting zone on the real environment, offering user feedback on the robot's progress and guiding efficient demonstration as shown in Fig.~\ref{fig:training}.

A modified version of this task was used in the transfer test for evaluating the user's performance in new tasks. The training task was basically flipped; the goal point became a starting point and the starting zone became the goal zone. This goal zone was shortened to a 4 x 8 test grid as shown in Fig.~\ref{fig:training}

\subsection{Performance Metrics}
\label{sec:metrics}
\begin{enumerate}
    \item \textbf{Task Completion Time:} We measured the total time, $t$, required by the user to complete the experiment task. 
    \item \textbf{Number of Demonstrations:} We measured the number of demonstrations $\lvert \mathcal{D} \lvert$ that achieve at least 90\% coverage of the starting zone.
    \item \textbf{Teaching Efficacy:} Efficacy refers to the ability to perform a task to a satisfactory or expected degree. Here, we used the teaching efficacy defined in~\cite{sena2020quantifying}, which is the ratio between the successful points,$|\hat{I}|$, and the total number of tested points, $|I|$, $(\epsilon = \frac{|\hat{I}|}{|I|})$.
     
    \item \textbf{Teaching Efficiency:} Efficiency in any given application is often context-dependent; however, typically it is desirable to minimize the total number of demonstrations required because this is correlated with both the time spent teaching and the space needed to store data. Here, we used the teaching efficiency defined in~\cite{sena2020quantifying}, which is the efficacy normalized by the number of the provided demonstrations, $\lvert \mathcal{D} \lvert$, 
    \begin{equation}
    \label{eq:efficiency}
        \eta = \frac{\epsilon}{\lvert \mathcal{D} \lvert}. 
    \end{equation}
\end{enumerate}




\subsection{Preliminary Experiment: Guidance Rule in Simulation}
\label{subsec:Grule}
In this experiment our goal is to compare the entropy-based guidance rule with the heuristic guidance rule proposed in~\cite{sena2020quantifying} (described in Section~\ref{sec:related_works}). Various factors influence robot learning and generalization, including the quality, quantity, and distribution of provided demonstrations. Here, we specifically focus on the guidance rule that governs the "number" and "distribution" of provided demonstrations, while keeping all other factors constant. To ensure consistent demonstration quality, we employed the motion planner RRT-Connect~\cite{kuffner2000rrt} to generate demonstrations for both guidance rules. Trajectories were generated from all points within the starting zone to the goal point and saved for the experiment. Additionally, to simulate a real-case scenario involving human demonstrations, we collected expert demonstrations, performed by one of the paper's authors, for all points within the starting zone of the experimental task shown in Fig.~\ref{fig:exp_setup}. 

For the comparison study, for each guidance rule, we developed a Python script to guide the robot's learning process. The experimental setup was simulated with RViz, the standard tool for visualization and interaction with robot applications implemented in ROS. The script initially proposes a point in the starting zone, and its corresponding saved trajectory (either from the motion planner or expert demonstrations) is retrieved. Then, the TP-GMM model is trained by this trajectory and tested with all points in the starting zone to identify the successful and failed trajectories. Using the guidance rule, the script selects the next point in the starting zone for the next demonstration. These steps are repeated until the learning model generates successful trajectories for 90\% or more of the points in the starting zone. The same procedure is repeated 56 times for both guidance rules until every point in the task space has been tested as the initial point. Finally, the two guidance rules are compared with respect to the number of demonstrations needed to generate successful trajectories for 90\% or more of the starting zone. 

\begin{figure}[t]
\centering
\includegraphics[width=0.48\textwidth]{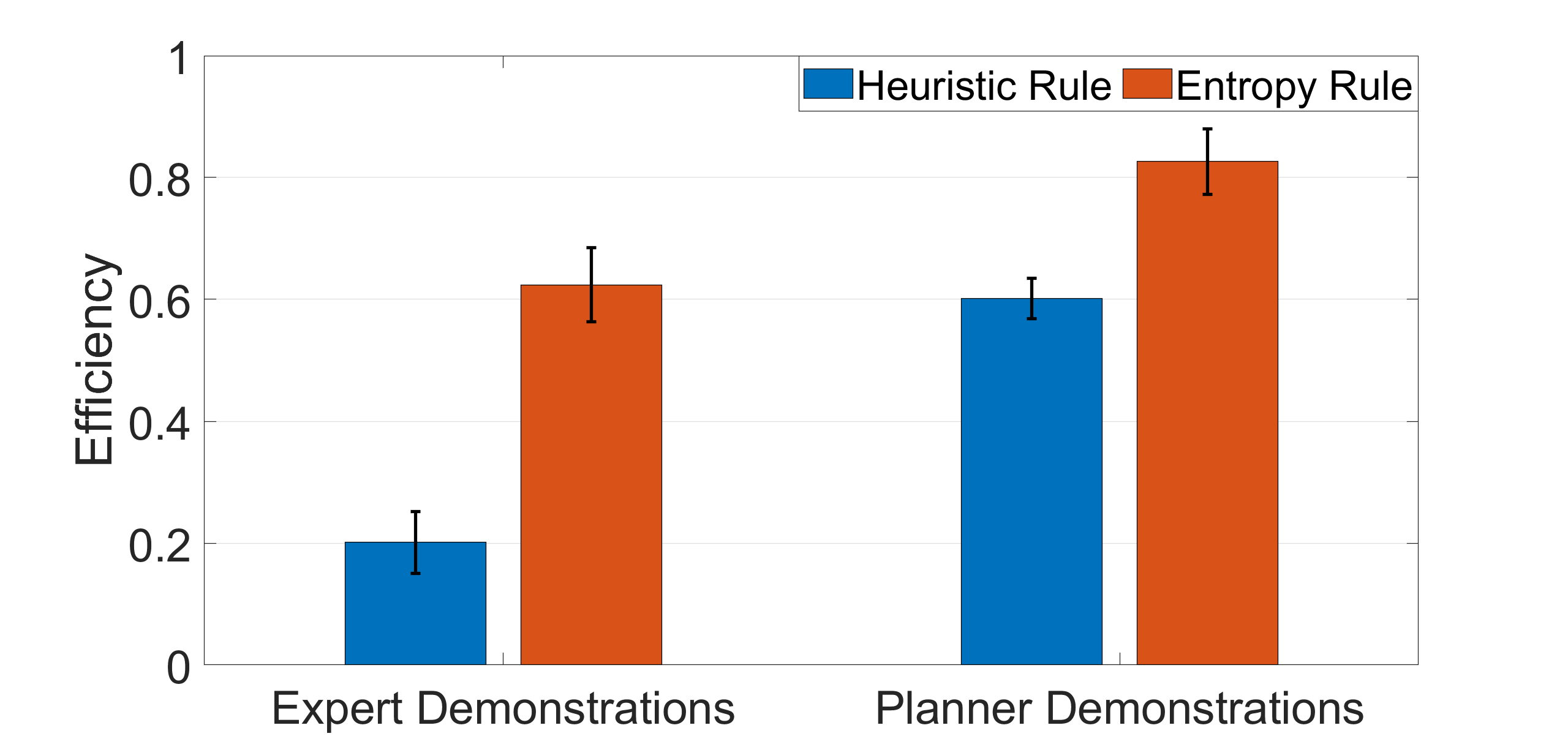}
\caption{The average robot learning efficiency for both the entropy rule and the heuristic rule driving the selection of planner and expert demonstrations.}
\vspace*{-3mm}
\label{fig:exp1_eff}
\end{figure}

\begin{figure*}
    \begin{subfigure}[b]{1\columnwidth}
         \centering
         \includegraphics[width=0.80\textwidth]{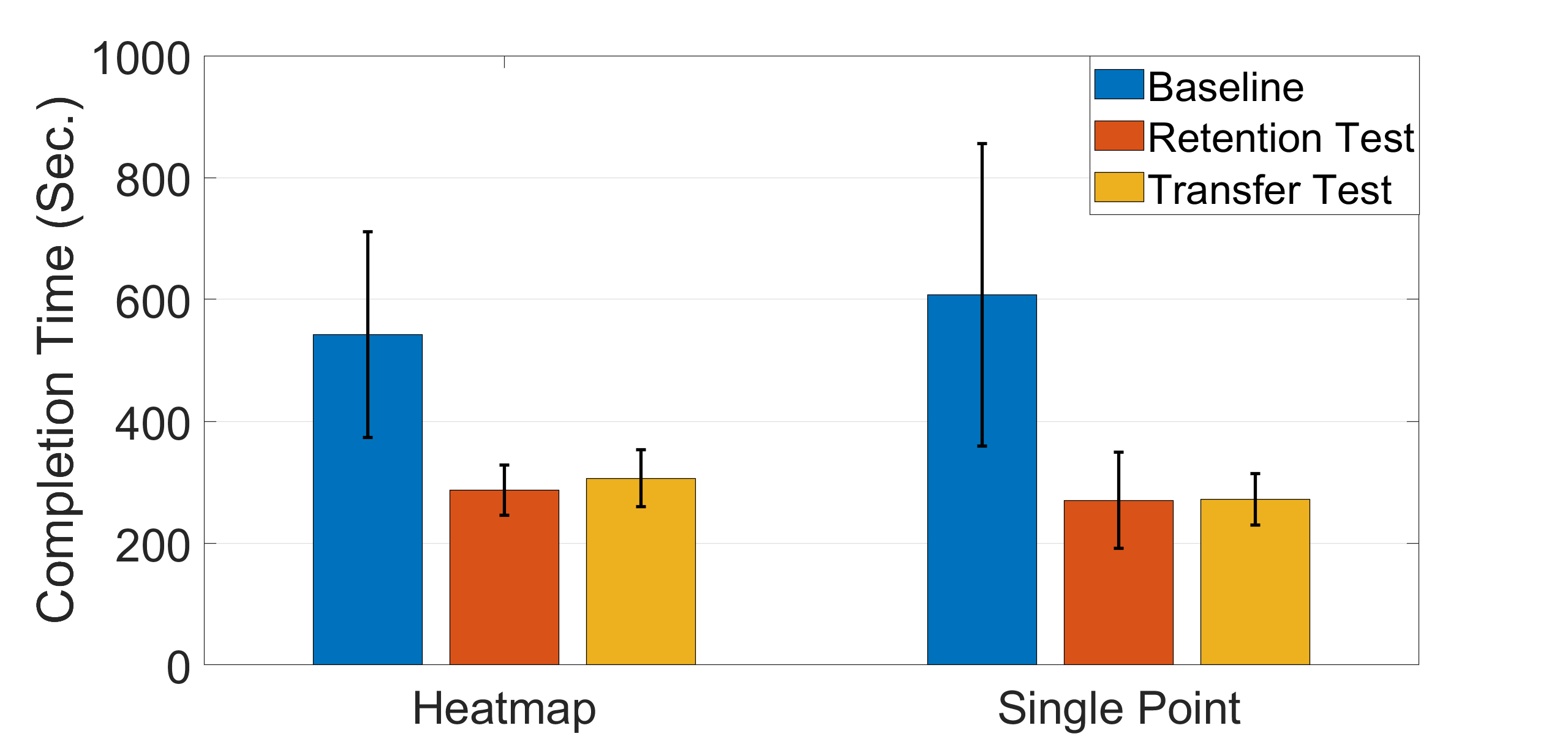}
     \end{subfigure}
     \hfill
     \begin{subfigure}[b]{1\columnwidth}
         \centering
         \includegraphics[width=0.80\textwidth]{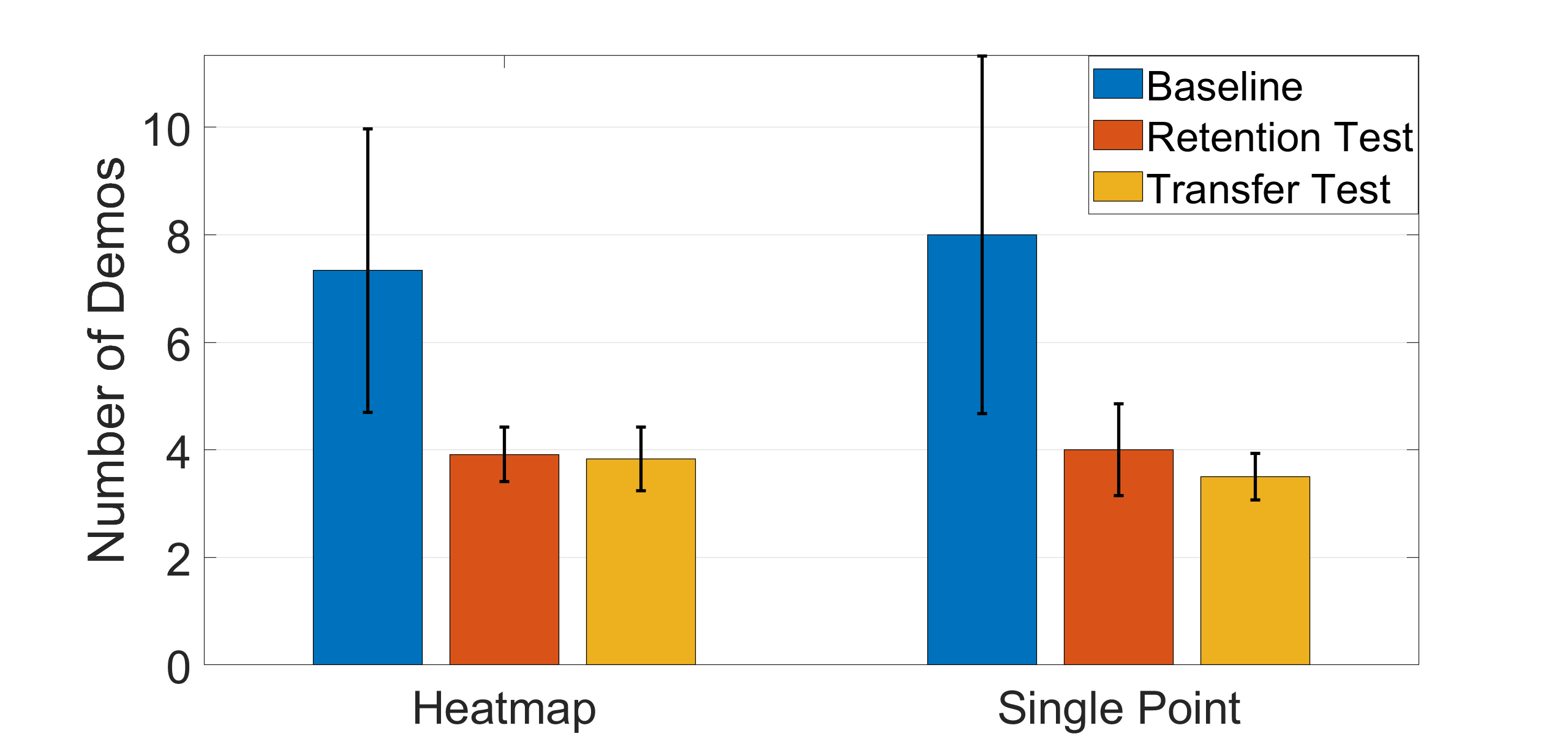}
     \end{subfigure}

     \begin{subfigure}[b]{1\columnwidth}
         \centering
         \includegraphics[width=0.80\textwidth]{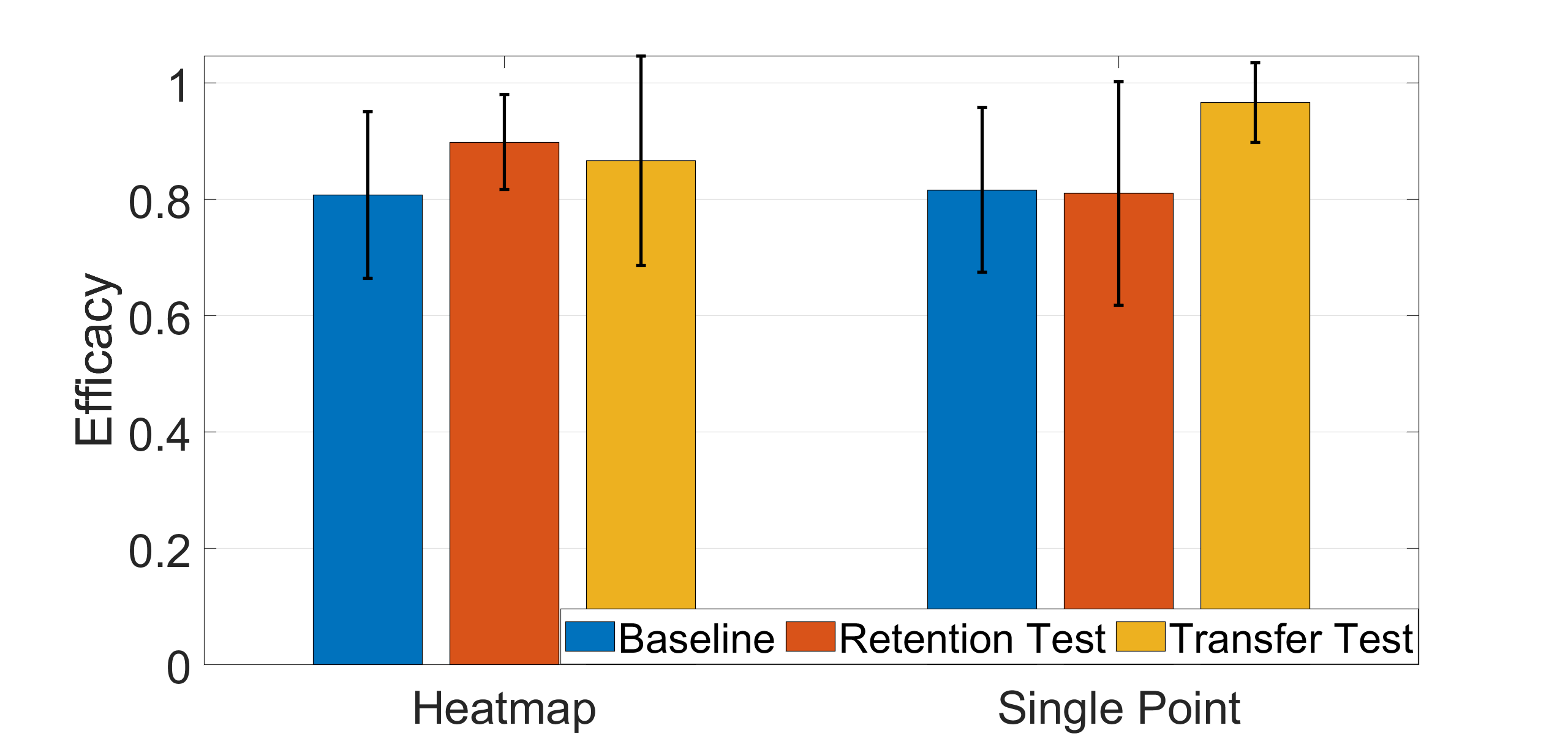}
     \end{subfigure}
     \hfill
     \begin{subfigure}[b]{1\columnwidth}
         \centering
         \includegraphics[width=0.80\textwidth]{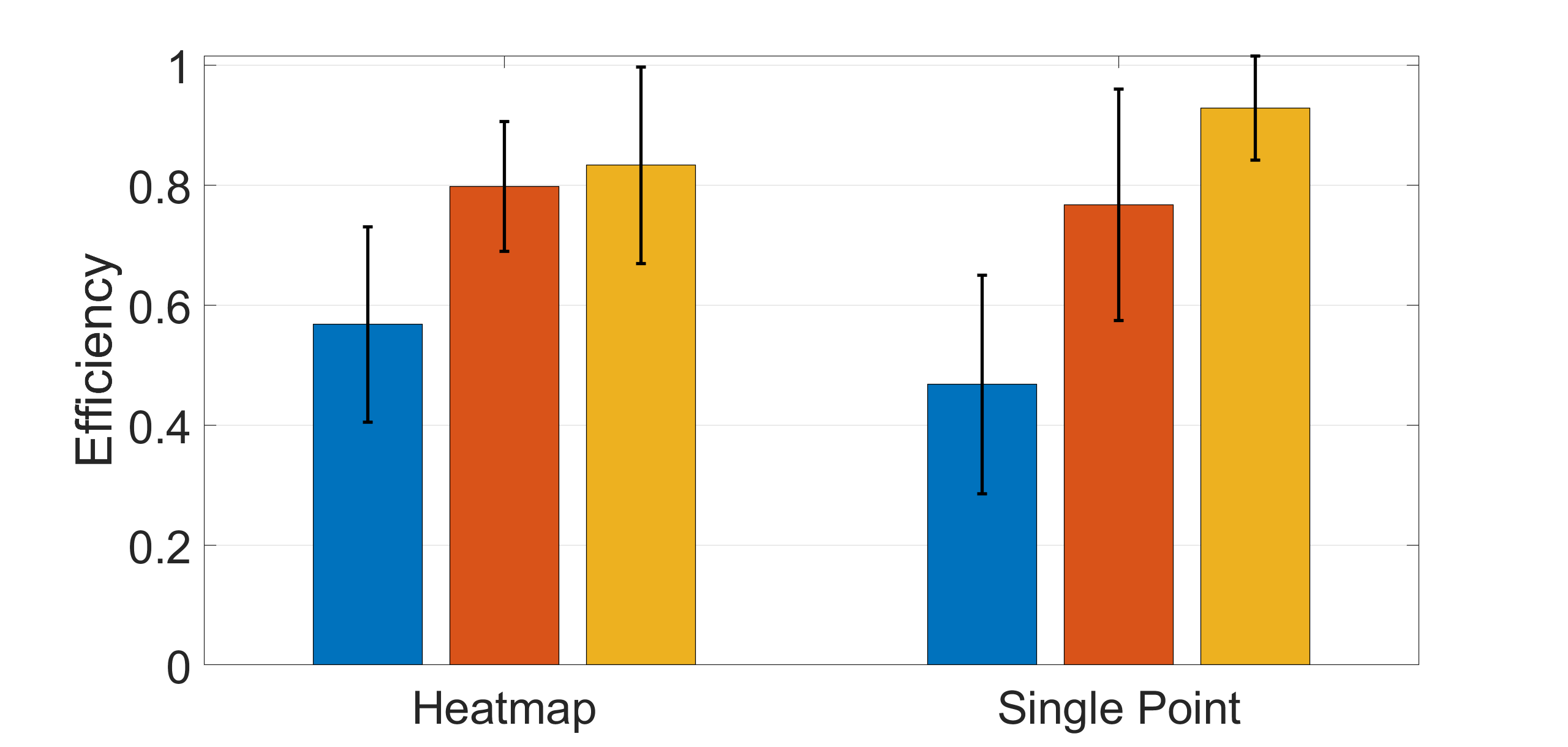}
     \end{subfigure}
        \caption{Performance metrics for comparing heatmap and single-point groups in baseline, retention and transfer tests}
        \label{fig:exp2_all}
\end{figure*}


\subsubsection{\textbf{Hypotheses}}~\\~
The experimental hypotheses are chosen to test whether robot learning efficiency, Equation ~\ref{eq:efficiency}, improves using the proposed approach compared to the heuristic rule. The alternative hypothesis is formally defined as:
\begin{itemize}
    \item\textbf{H1}: Robot learning efficiency using the entropy guidance rule is significantly higher than the efficiency using the heuristic rule using planner demonstrations.
    \item\textbf{H2}: Robot learning efficiency using the entropy guidance rule is significantly higher than the efficiency using the heuristic rule using human expert demonstrations.
\end{itemize}

\subsubsection{\textbf{Results}}~\\~
Figure~\ref{fig:exp1_eff} shows the mean and 95\% confidence interval of the robot learning efficiency in 56 trials using the heuristic rule and entropy rule with both expert and planner demonstrations. Overall, using the entropy rule achieves a higher robot learning efficiency than the heuristic rule, for both planner and expert demonstrations. A 2x2 Bayesian ANOVA was conducted to investigate the impact of a) guidance rule, and b) demonstration type on robot learning efficiency. Both the guidance rule $(BF_{10} = 7.43 \times 10^{15})$ and demonstration type $(BF_{10} = 2.82 \times 10^{13})$ show a large effect on robot learning efficiency. In addition, robot learning efficiency using either guidance rule strongly depends on the demonstration type (the quality of the provided demonstrations) $(BF_{10} = 184.11)$.  These results support \textbf{H1} and \textbf{H2}. 

Interestingly, a slight decrease in the demonstration’s quality (i.e., from planner to expert demonstrations) causes a drastic decrease in the robot learning efficiency by 200\% using the heuristic rule, while using the proposed entropy rule the efficiency decreased by only 33.9\%. This shows the importance of the guidance rule in robot learning from demonstrations and its contribution to the robot learning efficiency which cannot be compensated by only improving the demonstration quality. Thus, it is important to guide the users to provide an efficient set of demonstrations for the robot to learn from.

\subsection{User Study Experiment}
From the preliminary experiment results, we found that the entropy guidance rule significantly improves robot learning efficiency. So, we utilize this rule in a training framework for users to guide them to provide an efficient set of demonstrations for the robot. Figure~\ref{fig:training} shows an overview of the user study procedure as follows: 
\begin{enumerate}
    \item \textbf{Baseline:} The user provides as many demonstrations as they think would be sufficient for the robot to generate successful trajectories from any point in the starting zone to the goal position. The provided demonstrations are used to train a TP-GMM model and evaluate the performance to visualize feedback through the HoloLens to the user. The blue circles represent successfully learnt trajectories while the red circles represent failed learnt trajectories. 
    
    \item \textbf{Training:} The user provides an initial demonstration, and then the learning model is evaluated across the starting zone. The entropy is calculated in each area and visualized through the HoloLens as a heatmap or single-point, according to the training group. The user provides an additional demonstration in the highest entropy point and the entropy visualization is updated, and so on. This continues until the robot learns successful trajectories for 90\% or more of the starting zone. Each user was trained for three trials.
    
    \item \textbf{Retention Test:} This step is similar to the baseline to evaluate the performance after training.
    
    \item \textbf{Transfer Test:} The user teaches the robot to move from a specific starting point to a goal zone without guidance, described in Section~\ref{subsec:task}.
    
\end{enumerate}

The users were asked to fill in the NASA-TLX questionnaire~\cite{NASA-TLX} after their baseline and retention test to see the effect of the training on their perceived task load. They were also asked to fill in the SUS questionnaire~\cite{SUS} after the training to get their subjective evaluation of the proposed training system. Lastly, a semi-structured interview was conducted before and after training to inquire about users’ reasoning behind their selection of the number of provided demonstrations and their respective placements. These interviews were conducted with the intention of delving into users' mental models of robot learning before and after training.


\subsubsection{\textbf{Hypotheses}}~\\~
The experimental hypotheses are chosen to test whether the user's teaching skills measured through robot learning efficiency are significantly improved using the proposed approach. They are defined as:
\begin{itemize}
    \item\textbf{H3}: Robot learning efficiency will be significantly improved after training the human teacher using the proposed approach (over both groups).
    \item\textbf{H4}: The heatmap visualization group will achieve a significantly higher improvement in robot learning efficiency than the single-point visualization group on the retention test (i.e., same training task).
    \item\textbf{H5}: The heatmap visualization group will achieve a robot learning efficiency that is significantly higher than the single-point visualization group on the transfer test (i.e., on a novel task).
\end{itemize}

The user study is a between-participants design, so each participant is assigned to one group; either the heatmap group or the single-point group. A priori power analysis was conducted using G*Power version 3.1.9.7~\cite{faul2007g} to determine the minimum sample size required to test the study hypotheses. Results indicated the required sample size to achieve 80\% power for detecting a medium effect size (Cohen's \textit{f} = 0.3), and a type I error rate $\alpha = 0.05$, is $N = 24$, or 12 participants per condition. Accordingly, the results reported below are for a population of 24 participants (14 male, 10 female; ages $\mu = 24,\sigma = 4$). We recruited participants for our user study through advertisements posted on the University of British Columbia (UBC) campus and social media. Prior to conducting the study, we obtained research ethics approval from UBC’s Behavioural Research Ethics Board (application ID H20-03740-A001). We obtained informed consent from each participant before commencing the experiment.

\subsubsection{\textbf{Results}}~\\~
\label{subsec:res2}
\textbf{Quantitative Measures}~\\~
Figure~\ref{fig:exp2_all} depicts the mean and 95\% confidence interval for all performance metrics across baseline, retention, and transfer tests for both the heatmap and single-point groups. Overall, there is a noticeable improvement in all performance metrics during the retention and transfer tests when compared to the baseline performance. Efficiency improved by 140\% and 164\% from the baseline to the retention for both heatmap and single-point groups, respectively, and by 146\% and 198\% from the baseline to the transfer test for both heatmap and single-point groups, respectively. Interestingly, users' performance improved more in the transfer test than in the retention test compared to their baseline performance. Given that the efficiency metric encapsulates both the number of demonstrations and efficacy, the statistical analysis was centred on efficiency. 

Bayesian mixed analysis of variance (ANOVA) showed a strong impact of the training on efficiency when comparing the baseline to the retention test $(BF_{10} = 1387.16)$; supporting \textbf{H3}. The same analysis showed inconclusive results on the impact of the visualization type on efficiency $(BF_{10} = 0.4)$; no support for \textbf{H4} and \textbf{H5}. Furthermore, the Bayesian independent samples t-test showed inconclusive results on the comparison between the transfer test efficiency in both heatmap and single point groups $(BF_{10} = 0.60)$. This result highlights the effectiveness of the proposed entropy-based guidance system, irrespective of how the entropy information is visually presented to the user.

\begin{figure*}
    \begin{subfigure}[b]{1\columnwidth}
         \centering
         \includegraphics[width=0.85\textwidth]{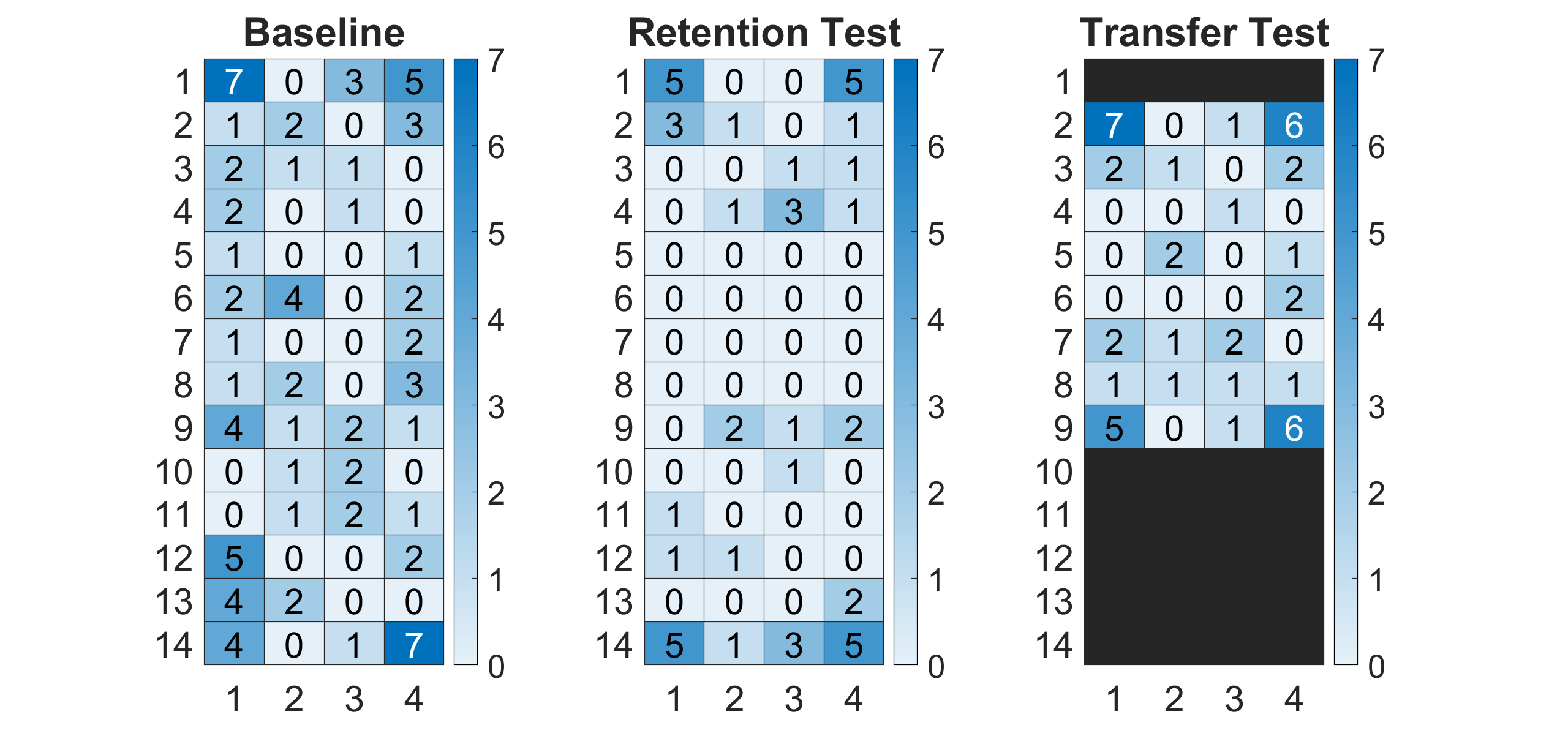}
         \caption{(a) heatmap group}
     \end{subfigure}
     \hfill
     \begin{subfigure}[b]{1\columnwidth}
         \centering
         \includegraphics[width=0.85\textwidth]{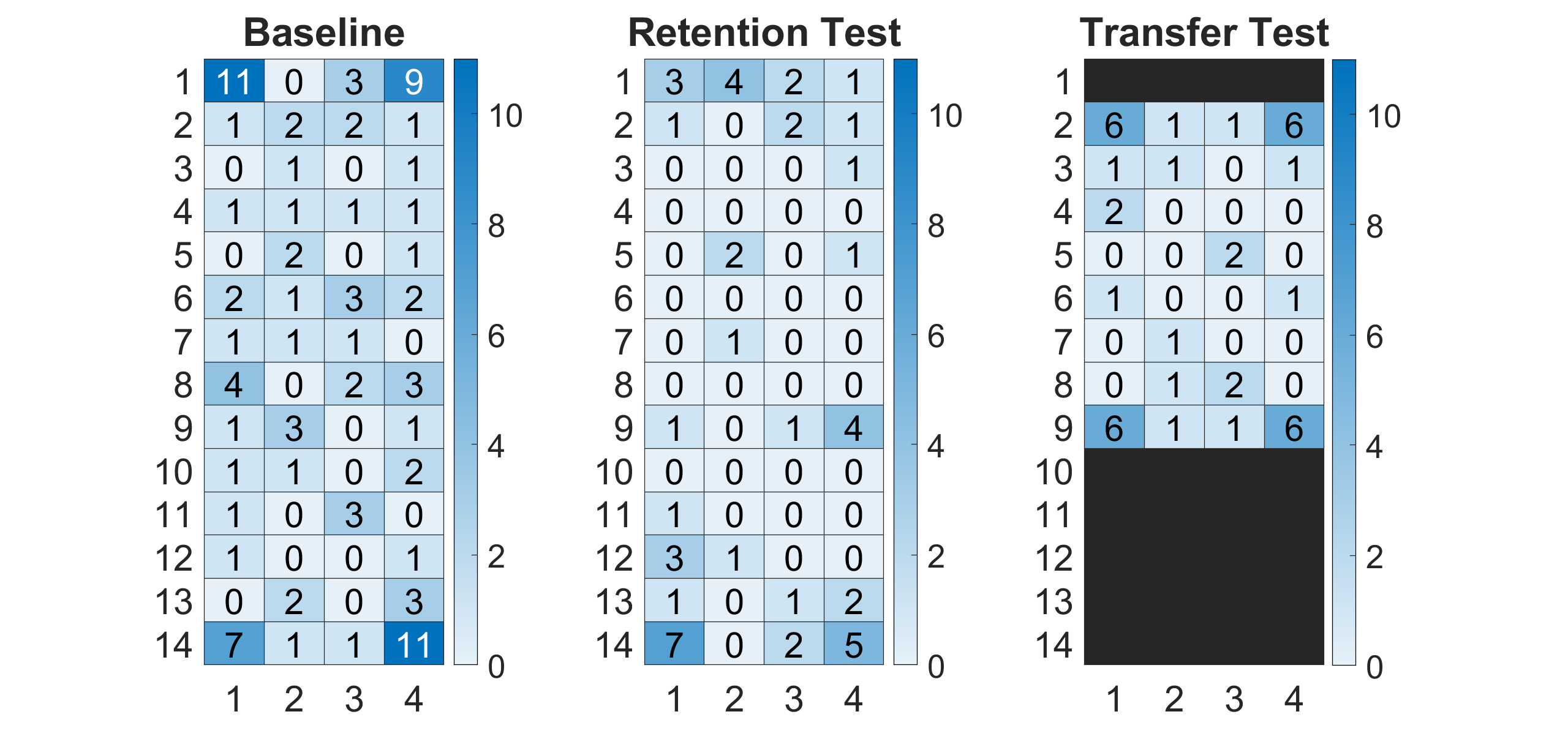}
          \caption{(b) single-point group}
     \end{subfigure}
     
     \caption{The number and distribution of the demonstrations in the starting zone for the baseline, retention, and transfer tests for both the (a) heatmap group and (b) single-point group. Users provided a high number of demonstrations and distributed them across the starting zone in the baseline. After training, they provided a lower number of demonstrations focused on the corners.}
     \label{fig:demos_freq}
\end{figure*}

We also investigated the distribution of the demonstrations over the task space across the baseline, retention and transfer tests. This shows the different strategies users adopted to distribute their demonstrations over the task space. Figure~\ref{fig:demos_freq} shows a heatmap representing the number and locations of the provided demonstrations across different stages of the experiment for both heatmap and single-point groups. The number of the provided demonstrations notably decreased from the baseline to the two tests. The median number of baseline demonstrations was 6 and 7, respectively, for the heatmap and single-point groups, decreasing to a median of 4 in the retention and transfer tests for both groups. Compared to the baseline which has a higher number of demonstrations from points in the interior of the start zone, the retention and transfer test demonstrations are concentrated at the corners. This highlights the change in users’ strategy for selecting the number and locations of demonstrations.

\begin{figure}[t]
\centering
\includegraphics[width=0.48\textwidth]{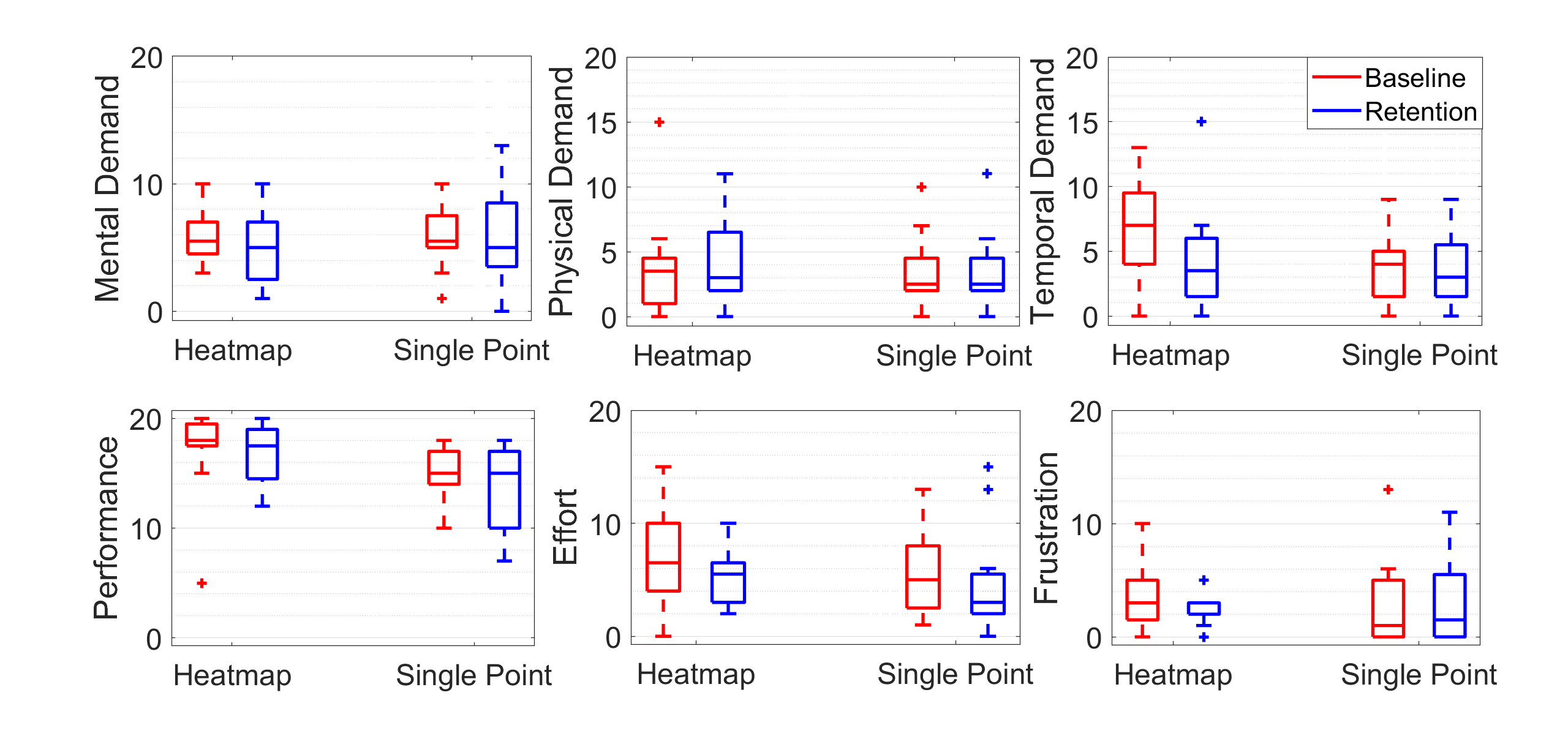}
\caption{Participants response to NASA-TLX \cite{NASA-TLX} questions after baseline and retention test. There is no significant difference before and after the training.}
\vspace*{-3mm}
\label{fig:nasa-tlx}
\end{figure}

\textbf{Qualitative Measures}~\\~
\textbf{Perceived Task Load:}
Figure.~\ref{fig:nasa-tlx} illustrates the box plots for both the heatmap and single-point groups following the baseline and retention tests. It shows that both effort and frustration have decreased after training in the heatmap group while only the effort decreased in the single-point group. In contrast to the quantitative measures, users' perceived performance after training was lower than before training in both groups. In general, the task load metrics statistically indicate no significant difference before and after the training. This suggests that the training did not have an adverse impact on users' perceived task load or lead to overwhelming experiences.


\textbf{System Usability:}
The usability score for the heatmap group is (M = 75.21; 95\% CI, 64.99  to  85.42), while in the single point group (M = 62.71; 95\% CI, 55.89  to  69.53). According to the global benchmark for SUS created by Sauro and Lewis through surveying 446 studies spanning different types of systems, the mean given score is $68 \pm 12.5$ \cite{Sauro2012}. When comparing this global benchmark mean score with the scores obtained from the two training groups, the observed differences were inconclusive (Bayesian one-sample t-test $BF_{10} = 0.75, BF_{10} = 0.89$, respectively). This suggests that both training approaches are usable from the user’s perspective.

\textbf{Interviews:}
Participants employed diverse strategies when determining the number and placements of demonstrations during the baseline phase. Among the 24 participants, 19 opted to establish a grid of points, strategically encompassing both the corners and intermediary positions. As articulated by one participant: \textit{"Well, overall, I just wanted to get the corners and then a few different points along the diagonals in between. I just felt like it kind of covers the majority of the grid"} which is reflected in Fig.~\ref{fig:demos_freq}. In contrast, two other participants opted for a completely random selection of demonstrations, favoring a higher count of demonstrations (18 and 23) to ensure comprehensive coverage of the starting area. Another participant thought that providing a dense set of points would allow the robot to learn the distance between the points in the starting zone and generalize well to other areas. 

After training, almost all participants provided an average number of demonstrations in retention equal to the average number of demonstrations they provided in the training, regardless of the training condition. As one participant said, \textit{"After the second training trial, I realized that I don't need a really large number of points as long as I pick the four corners of the rectangular area"}. In addition, there was an agreement from several participants in the heatmap group that visualizing the generalization area of each point in the training was very helpful in deciding the location of the next demonstration in the retention test. Some participants counted the number of generalized points around one demonstration in the training and tried to distance the demonstrations in the retention by that number of points.



\section{DISCUSSION}
        \label{sec:discussion}
        This study demonstrated that a few trials of interactive training and guidance for lay users significantly improved their teaching skills, which in turn improved robot learning and generalization efficiency. It's noteworthy that this learning and generalization happened online, using demonstrations from a teacher with no prior knowledge of robotics or machine learning algorithms. By using the proposed training framework, users develop an understanding of what demonstrations the robot requires to learn efficiently, without necessarily understanding how learning is taking place. Thus, the overarching goal of this work is realized by decreasing the teaching cost by providing a small number of demonstrations that achieve wider generalization.


From the user study results and statistical analysis, there is support for \textbf{H3} and no support for \textbf{H4} or \textbf{H5}. Both training groups showed a significant improvement in learning efficiency from the baseline to the retention and transfer tests. This underscores the importance of training novice users before teaching the robot, which is consistent with prior work ~\cite{cakmak2014teaching, calinon2007teacher}. In addition, this study shows the benefit of the proposed entropy guidance system regardless of the visualization used in the training. Notably, both training groups achieved similar performance in retention and transfer tests. This is because most of the participants have a preconception of creating a grid of points to cover the starting zone to help the robot learn better, as shown in Fig.~\ref{fig:demos_freq}. Subsequently, post-training, participants recognized the feasibility of achieving high efficiency with a reduced number of demonstrations, contrary to their initial assumptions. As shown in Fig.~\ref{fig:exp2_all}, the efficacy before and after training is similar. This finding is unlike previous work~\cite{sena2020quantifying} that found most of the participants tend to underestimate the number of demonstrations required to teach the robot effectively.

A surprising observation emerged concerning users' perceived performance post-training. Despite a significant quantitative improvement in performance, 8 out of the 24 participants reported a decline in perceived performance compared to before training. This discrepancy was investigated by analyzing their interview data related to their approach to determining the number and locations of demonstrations. These participants experienced significant shifts in their strategies after training, which impacted their confidence and perception of performance. For instance, one participant provided 16 demonstrations in the baseline. After training they provided four demonstrations in the retention test and remarked in the interview \textit{“I just used four points, although I'm not sure I covered 90\%, I don't know”.} Others changed their strategies from dense point sets to a more distributed approach, albeit with scepticism. These findings suggest the potential need for adaptive training tailored to individual participants' requirements, especially for those with strong pre-training strategy convictions, aligning with previous research~\cite{9832744}.

Another interesting observation is that one participant without prior knowledge in robotics realized the importance of the demonstration's quality in conjunction with the number and location of the provided demonstrations for robot learning and generalization. They tried different starting points in the training, and then they realized that providing smooth demonstrations without abrupt corners allows the robot to learn and generalize faster with a smaller number of demonstrations. However, it's challenging to extrapolate that all participants would naturally arrive at this realization without proper training. This underscores the necessity for an instructional framework to guide participants in offering demonstrations of both high quality and optimal efficiency, echoing the emphasis of previous research~\cite{argall2009survey}.

\section{CONCLUSION AND FUTURE WORK}
        \label{sec:conclusion}
        Toward our goal of achieving efficient robot learning and generalization, we proposed using information entropy as a criterion for guiding human teachers to provide the next demonstration that achieves the highest efficiency. This proposed approach reduces the teaching dimension by suggesting the minimum number of demonstrations that achieves the highest learning efficiency. The proposed approach was validated in two experiments to explore its contribution to both efficient robot learning and generalization as well as improving novices' teaching skills. We found that the proposed approach significantly improves robot learning efficiency compared to the state-of-the-art heuristic rule. 

We believe that the results of this paper open up several directions for future research. The significant improvement in the efficiency by only guiding users to well distribute the demonstrations suggests that guiding users to provide high-quality demonstrations as in~\cite{sakr2020training} along with their good distribution could further boost learning efficiency. It will also be interesting to test the proposed approach in facilities with real users without controlling the conditions. For instance, users would then have the freedom to decide how long they need guidance to ensure that they provide the most beneficial demonstrations to the robot.


\bibliographystyle{IEEEtran}
\bibliography{bibliography}

\end{document}